\documentclass{article}
\usepackage{subcaption}
\usepackage{subcaption}
\usepackage{authblk}
\usepackage{amsmath}
\usepackage{graphicx} 
\usepackage{adjustbox} 
\usepackage{placeins}
\usepackage{float}
\usepackage{caption} 
\usepackage{lipsum} 
\usepackage{array}
\title{Performance Evaluation of YOLOv8 Model Configurations, for Instance Segmentation of Strawberry Fruit Development Stages in an Open Field Environment}
\author[1]{Abdul-Razak Alhassan Gamani}
\affil[1]{Department of Agricultural and Biosystems Engineering, KNUST}
\author[2]{Ibrahim Arhin}
\affil[2]{Department of Agricultural and Biosystems Engineering, KNUST}
\author[3]{Adrena Kyeremateng Asamoah}
\affil[3]{Faculty of Agriculture, KNUST}

\date{July 2024}

\usepackage{graphicx}
\usepackage{placeins}
\begin{document}

\maketitle
\begin{abstract}
Accurate identification of strawberries during their maturing stages is crucial for optimizing yield management, and pest control, and making informed decisions related to harvest and post-harvest logistics. This study evaluates the performance of YOLOv8 model configurations for instance segmentation of strawberries into ripe and unripe stages in an open field environment. 
The YOLOv8n model demonstrated superior segmentation accuracy with a mean Average Precision (mAP) of 80.9\%, outperforming other YOLOv8 configurations. In terms of inference speed, YOLOv8n processed images at 12.9 milliseconds, while YOLOv8s, the least-performing model, processed at 22.2 milliseconds.
Over 86 test images with 348 ground truth labels, YOLOv8n detected 235 ripe fruit classes and 51 unripe fruit classes out of 251 ground truth ripe fruits and 97 unripe ground truth labels, respectively. In comparison, YOLOv8s detected 204 ripe fruits and 37 unripe fruits.
Overall, YOLOv8n achieved the fastest inference speed of 24.2 milliseconds, outperforming YOLOv8s, YOLOv8m, YOLOv8l, and YOLOv8x, which processed images at 33.0 milliseconds, 44.3 milliseconds, 53.6 milliseconds, and 62.5 milliseconds, respectively.
These results underscore the potential of advanced object segmentation algorithms to address complex visual recognition tasks in open-field agriculture effectively to address complex visual recognition tasks in open-field agriculture effectively.

\end{abstract}

\section{Introduction}
Instance segmentation using deep learning involves training algorithms to automatically identify and segment objects within digital images, leveraging large datasets and complex neural networks to learn distinguishing features \cite{Mittal2019} \bibliographystyle{ieeetr}. In the agriculture industry, this technology is pivotal for automating tasks such as pest detection, fruit counting, and crop monitoring, enhancing precision and efficiency \cite{Tian2020},\cite{Delnevo2022}. By enabling accurate and rapid analysis of vast amounts of visual data, deep learning-based object detection significantly contributes to sustainable farming practices, optimized resource use, and increased crop yield \cite{Sharma2021}.

Strawberry is a fruit that is high in essential phytochemicals and nutrients, which are very beneficial to the human body \cite{giampieri2012strawberry}.
In the domain of precision and automated agriculture, the integration of machine vision for fruit detection enhances the efficiency of robotic harvesting systems \cite{li2011review}. These technologies are pivotal for greenhouse and open-field growers, aiding in the precise prediction of fruit quality, market pricing, and the management of harvest and post-harvest processes \cite{Stajnko2004}. During fruit growth, the detection of irregular growth traits such as variations in size and shape can indicate issues like pest infestations or nutrient deficiencies \cite{pest2024}, \cite{Campos2021}. Moreover, the rapid growth rate of strawberries and the labor-intensive nature of manual harvesting underscore the importance of automation in maintaining high yields and preventing the rapid deterioration of strawberry plants after production \cite{Wu2019}, \cite{Erfani2019}.
However, the field of deep learning is continuously evolving, with the development of newer models that offer enhanced capabilities and performance. This constant advance in technology brings sophisticated tools to the forefront of agricultural applications. Among these, the YOLO (You Only Look Once) algorithm stands out as a particularly influential development in the field of agricultural automation. Originally designed for real-time object detection, YOLO has been extensively adapted for various agricultural purposes, including the specific challenges associated with strawberry detection using instance segmentation. 
Recent adaptations of YOLO models have significantly improved the accuracy and efficiency of detecting strawberries amidst complex backgrounds where they camouflage with foliage. These improvements are crucial given the unique challenges posed by strawberries' physical characteristics. As a testament to its growing utility, a variety of YOLO-based systems have been implemented to automate strawberry detection over the past few years. 

\section{Related Literature}
In determining a good model use case, existing literature on object detection, instance segmentation, yield analysis, and monitoring of fruit development of strawberries and other fruits were reviewed. Strawberry has a rapid growth rate; hence automation is necessary for monitoring growth, predicting yield analysis, and harvesting operations. Different authors have utilized distinguished methods for the recognition and detection of fruits by applying spectrum analysis, machine learning, and multi-template matching algorithms. A study by \cite{yuan2011greenhouse}  used principal component analysis (PCA) to perform dimensionality reduction on the near-infrared (780-880)nm and visible-light (650-700)nm bands, and then input the reduced-dimensional features into the backpropagation (BP) neural network for cucumber recognition. This approach is efficient in cucumber recognition. However, there are limitations in spectral analysis such as the need for additional light sources and the large size of acquired image data, making image analysis time-consuming. In contrast, the machine learning method facilitates practical, fast, and interesting data analysis in precision agriculture. 
The YOLO series models have also gained significant attention for their excellent role in the field of object detection and segmentation.\cite{Liu2020} introduced a modified version of YOLOv3, called YOLO-Tomato (designed to address lighting changes, over-lapping, and occlusions). The proposed model used circular bounding boxes, achieving a mean Average Precision (mAP) of  94.58\%  under challenging conditions using a dataset containing 609 images. \cite{Wang2021} provided an improved version of this model by incorporating K-means clustering to improve the box size calculation and facilitate multi-scale training. The experimental result presents a significant improvement in the mAP, achieving an overall value of 96.41\%. \cite{ruparelia2022real} compared YOLOv3 and YOLOv4 over a dataset containing 2000 images. The results indicated that YOLOv3 achieved a lower mAP of 78.49\%, while YOLOv4 achieved mAP of 81.28\%, highlighting the advantage of denser models in achieving improved precision in phenotypic trait detection. YOLOv5 has garnered attention for its impressive performance in terms of accuracy and speed. \cite{Wang2023} proposed the detection of small targets such as apple fruits using YOLOv5s and by applying improvements to the RFA module, DFP module, and Soft-NMS algorithm. The experimental results present an encouraging performance that the integration of the improved model achieved a significant improvement in detection accuracy, with precision, recall, and mAP increasing by 3.6\%, 6.8\%, and 6.1\%, respectively. proposed a method where he combined the flexible and efficient training tools with a proposed architecture and the compound scaling method. Results indicated that the YOLOv7 was better than all known detectors in both speed and accuracy proposed an improvement to the YOLOv8 framework characteristically tailored for tomato harvesting automation, implementing a quality improvement model to enhance feature extraction, replacing deeply distinguishable convolution with regular convolution to reduce computational complexity, and introducing a two-way attention gate for enhancing the overall recognition accuracy. The experimental result led to an overall mAP of 93.4\% on the proposed dataset. 
A methodology, for instance, segmentation of strawberries using deep learning techniques was proposed by \cite{Borrero2020} based on Mask R-CNN with a reduced architecture for the backbone and the Mask network. The study also proposed a new performance metric called the Instance Intersection Over Union (I2oU), to assess different options of instance segmentation techniques. Another study by \cite{Afzaal2021} instance segmentation model based on Mask R-CNN for detecting seven different types of strawberry diseases. The final model achieved a mAP of 82.43\% on the test set, outperforming other object detection architectures like YOLACT.
A lightweight YOLOv5$-$LiNet model for fruit instance segmentation by \cite{Lawal2023} was proposed to address the challenges of complex and changing environments, speed, accuracy, and lightweight requirements for low-power computing platforms. The model uses a modified YOLOv5n backbone with Stem, Shuffle\_{Block}, ResNet, and SPPF networks, a PANet neck network, and an EIoU loss function to improve detection performance. A study by \cite{Yang2023} investigated the LW$-$Swin Transformer for strawberry ripeness detection for efficient segmentation alongside the YOLOv8 algorithm whilst implementing the 5$-$fold cross$-$validation for a more comprehensive model evaluation. 
Another paper presented by \cite{Jia2022} used the YOLOF-Snake model, an efficient segmentation model for green object fruit detection and segmentation. The model used a single-layer feature map from the ResNet101 backbone network and an encoder-decoder structure to detect and classify the green fruit. In the study of \cite{Chai2023}, the article discusses the development and implementation of a real-time strawberry ripeness detection system using augmented reality (AR) and deep learning. The researchers used the YOLOv7 deep learning model for object detection and classification of strawberry ripeness levels (unripe, partially ripe, ripe) exploring the use of multi-scale training and a lightweight YOLOv7-tiny model to balance detection accuracy and speed. \cite{Li2024} utilized the YOLOv7 object detection algorithm and RGB-D sensing for strawberry recognition and positioning for robotic harvesting. A proposal by \cite{Yue2023} utilized an improved YOLOv8s-Seg algorithm for segmenting healthy and diseased tomatoes in the growth stage. 
A study by \cite{Yu2019} proposed a strawberry fruit detection algorithm based on Mask R-CNN (MRSD) for harvesting robots, which overcame the difficulties of poor universality and robustness using traditional machine vision algorithms in a non-structural environment.  Instance segmentation image output from MRSD provided a powerful basis for locating the picking point of strawberry fruit, which is convenient for the precise operation of the harvesting robot. In minimizing labor during strawberry production and harvest \cite{Yamamoto} developed a strawberry harvesting robot that can operate on a hanging bench culture. In quantifying the number of fruit detections per image, \cite{Gonzalez2019} proposed a network based on Mask R-CNN for object detection and instance segmentation for quantification of blueberry fruits. 
As can be observed from the reviewed literature, at the time of writing, there was a minimal focus on evaluating the performances of the various YOLO series model configurations in yield analysis for strawberry fruits, leaving questions on the most effective model to use for potential accuracy unanswered. The main objective of this study was to evaluate the performance of the YOLO series models for predicting ripe and unripe strawberry fruits during their growth stages. In this paper, the YOLO series models are used in the segmentation of fruit images of open-field strawberry plants. Using these YOLO series models allows for rapid detection of specific images irrespective of the color similarities. The best algorithm will be determined based on detection performance metrics such as accuracy and efficiency. By accurately analyzing and detecting the quantity of fruit, we can more precisely estimate the yield of individual strawberry fruits and monitor their growth. For real-time detection on edge devices\cite{lemsalu2022real} utilized the YOLOv5 model for strawberries and its peduncle detection for the robotic picking system in an open strawberry field. 

\section{Materials and Methods}
The study consisted of three major stages, as shown in Figure 1, beginning with acquiring RGB data in an open field environment, then ripe and unripe strawberry fruit segmentation, and finally fruit yield estimation. The dataset was acquired from Supervisely Dataset Ninja, an online platform for hosting open computer vision datasets. By employing deep learning methodologies, image data was trained on five state-of-the-art YOLOv8 configuration models. The dataset was grouped into three sets; training test, validation set, and test set by the ratio of 80:10:10. Strawberry fruit yield estimation was achieved by counting the number of detections per class for every image in the test set directory by running inference through Roboflow model API.

\begin{figure*}[ht!]
    \centering
\includegraphics[width=0.8\linewidth]{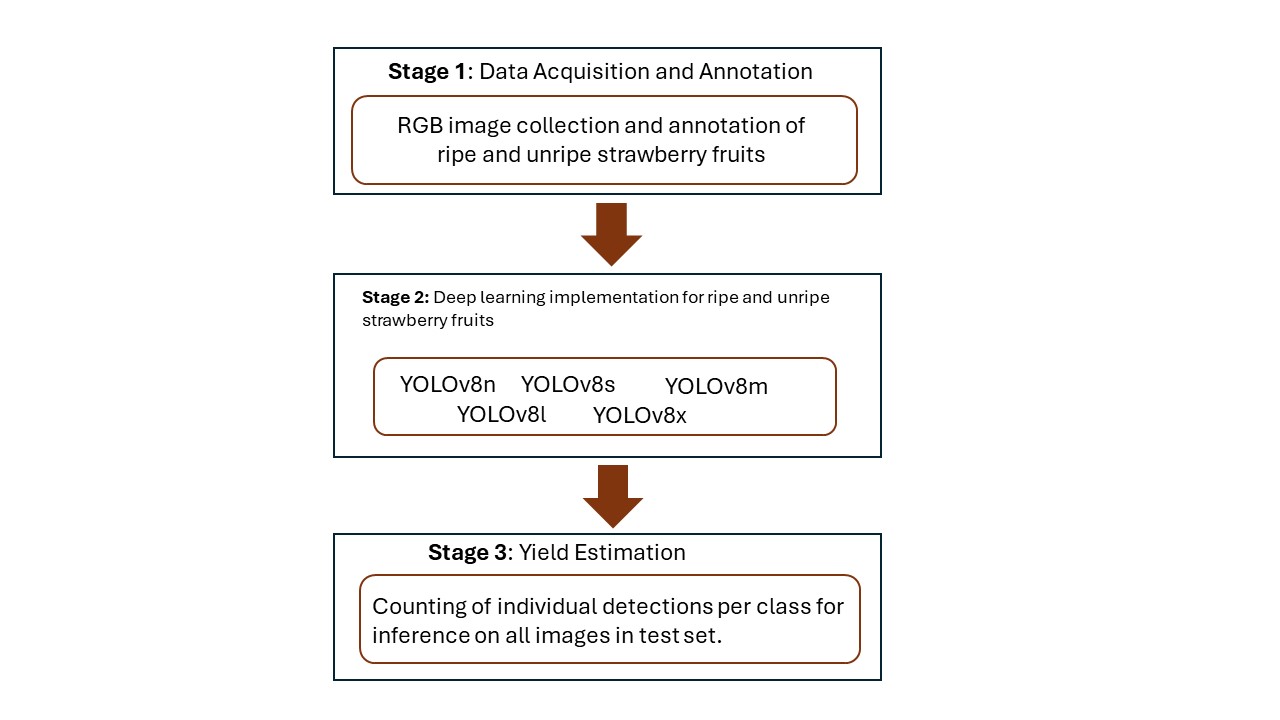}
    \caption{Block diagram illustrating the strawberry instance segmentation workflow and yield estimation}
    \label{fig: Figure 1}
\end{figure*}
\subsection{Data Acquisition and Annotation}
The dataset, released in 2022 by the Natural Resources Institute Finland (Luke) and hosted on Supervisely Dataset Ninja, comprises 813 images with 4568 labeled objects across three classes: ripe, peduncle, and unripe, for open-field strawberry detection. For instance segmentation, the dataset was augmented to 1386 training, 87 validation, and 86 test images, grouped into two classes: ripe and unripe strawberries (Figure 2 and Figure 3).

\begin{figure*}[!ht]
    \centering
    \begin{minipage}{0.49\textwidth}
    \centering
    \includegraphics[width=0.8\linewidth]{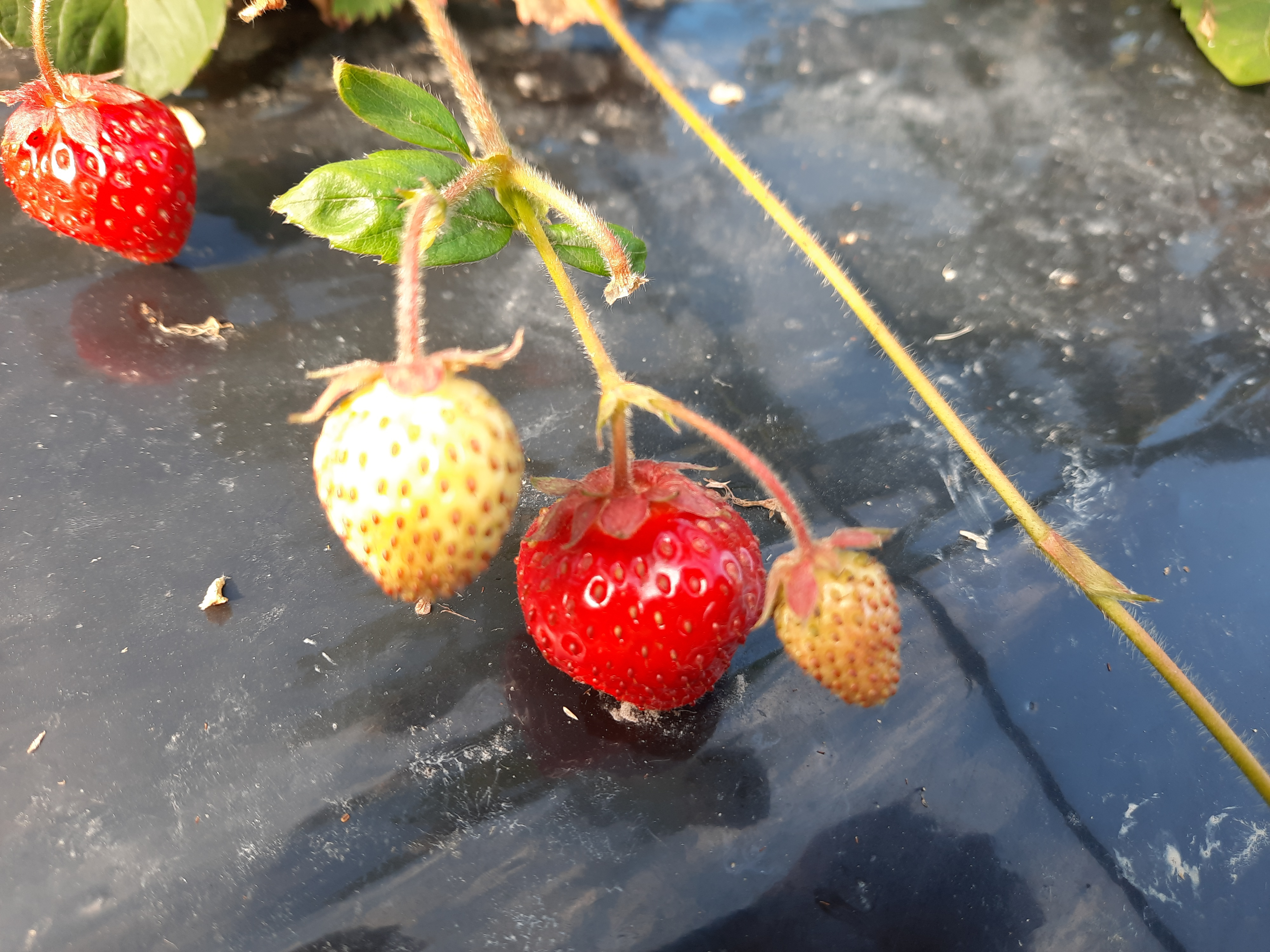}
    \caption{ripe and unripe fruits}
    \label{fig:2a}
    \end{minipage}
    \hfill
    \begin{minipage}{0.49\textwidth}
    \centering
    \includegraphics[width=0.8\linewidth]{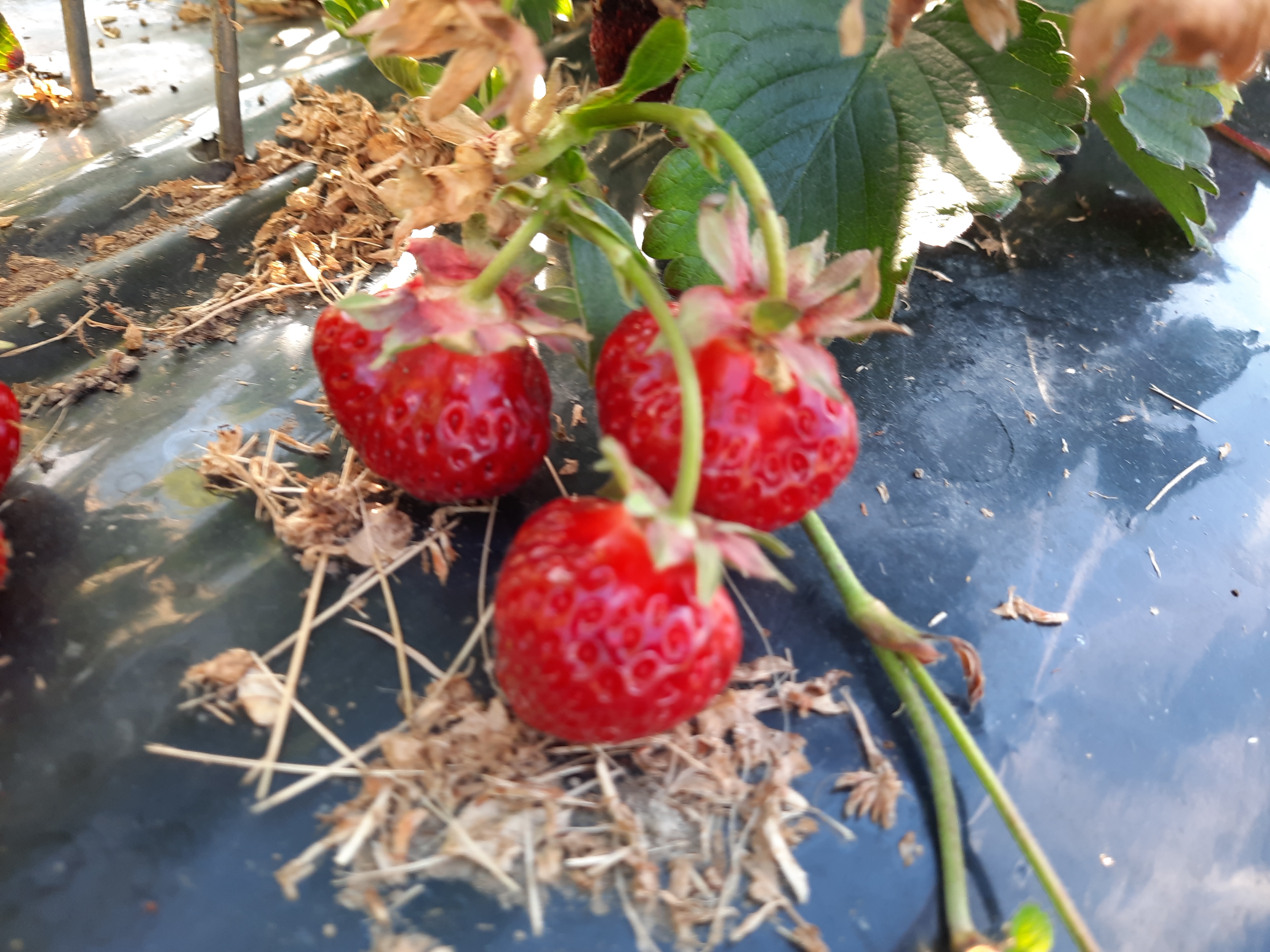}
    \caption{ripe fruits}
    \label{fig:2b}
    \end{minipage}
\end{figure*}
The Roboflow annotation tool was utilized to label the target objects as ripe and unripe strawberry fruits, as shown in Figure 4 and Figure 5.

\begin{figure}[!ht]
    \centering
    \begin{minipage}{0.49\textwidth}
    \centering
    \includegraphics[width=0.8\linewidth]{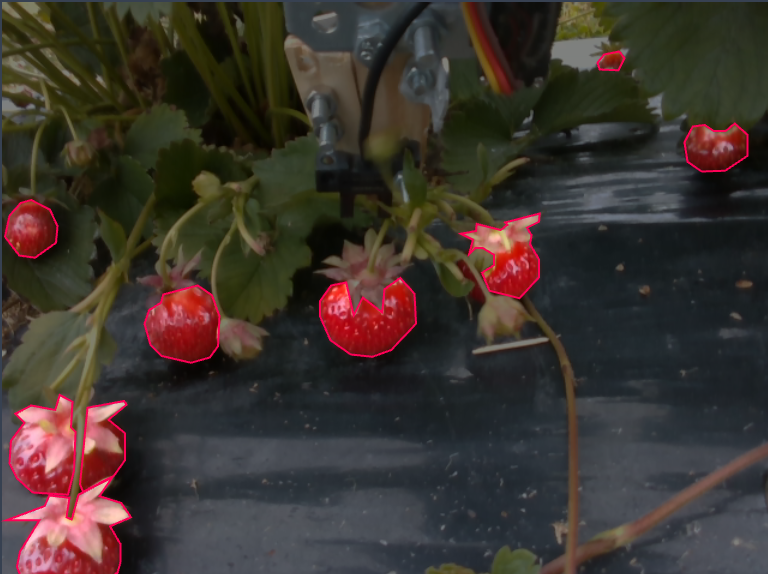}
    \caption{Annotation for unripe fruits}
    \label{fig: Annotation for unripe fruits}
    \end{minipage}
    \hfill
    \begin{minipage}{0.49\textwidth}
    \centering
    \includegraphics[width=0.8\linewidth]{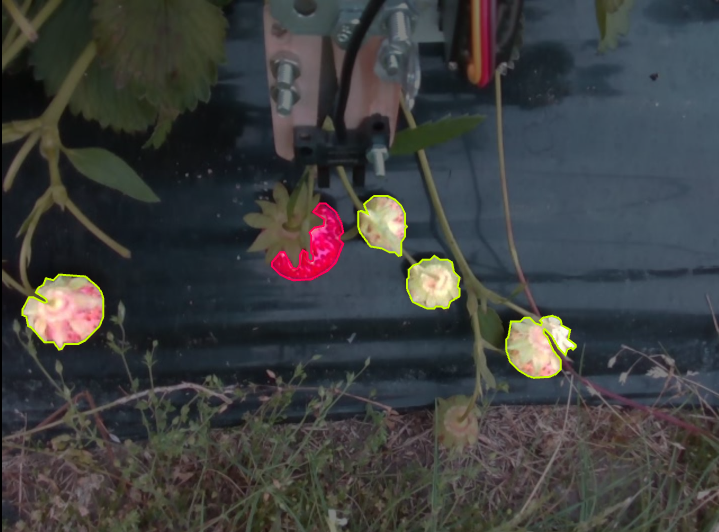}
    \caption{Annotation for unripe fruits}
    \label{fig: Annotation for unripe fruits 2}
    \end{minipage}
\end{figure}

\section{Training}
This study utilized the state-of-the-art YOLOv8 model. All models were trained on Google Colaboratory for Research using the Persistence-M Nvidia GPU (T4).
\subsection{Model Training and Testing}
This research utilized specific YOLOv8 model configurations: YOLOv8n, YOLOv8s, YOLOv8m, YOLOv8l, and YOLOv8x instance segmentation models recently provided by Ultralytics. Several training parameters were set for all models, including a batch size of 16, a learning rate of 0.01, 100 training epochs, a Stochastic Gradient Descent for optimization, an input size of 640x640 pixels, and a patience of 30 epochs per training.
\subsubsection{YOLOv8n}
As mentioned earlier, the YOLOv8n model configuration was trained for 100 epochs with early stoppage patience set at 30 epochs if validation loss did not improve. The training was completed at the 100th epoch within 2.561 hours. A batch size of 16 and an image size of 640 × 640 pixels were utilized. The YOLOv8n model consists of 195 layers, 3,258,454 parameters, 0 gradients, and 12.0 GFLOPs.
\subsubsection{YOLOv8s}
The YOLOv8s model configuration was trained for 100 epochs with early stoppage patience set at 30 epochs if validation loss did not improve. Training was halted at epoch 94, as no improvement was observed in the last 30 epochs, with the best results at epoch 64. The training duration was 2.755 hours. A batch size of 16 and an image size of 640 × 640 pixels were used. The YOLOv8s model consisted of 195 layers, 11,780,374 parameters, 0 gradients, and 42.4 GFLOPs.
\subsubsection{YOLOv8m}
The YOLOv8m model configuration was also trained for 100 epochs with early stoppage patience set at 30 epochs if validation loss did not improve. Training completed at the 100th epoch within 2.994 hours. A batch size of 16 and an image size of 640 × 640 pixels were used. The YOLOv8m model consisted of 245 layers, 27,223,542 parameters, 0 gradients, and 110.0 GFLOPs.
\subsubsection{YOLOv8l}
The YOLOv8l model configuration was trained for 100 epochs with early stoppage patience set at 30 epochs if validation loss did not improve. Training was halted at epoch 95, as no improvement was observed in the last 30 epochs, with the best results at epoch 65. The training duration was 3.205 hours. A batch size of 16 and an image size of 640 × 640 pixels were used. The YOLOv8l model consisted of 295 layers, 45,913,430 parameters, 0 gradients, and 220.1 GFLOPs.
\subsubsection{YOLOv8x}
The YOLOv8x model was trained for 100 epochs with early stoppage patience set at 30 epochs if validation loss did not improve. Training was halted at epoch 43, as no improvement was observed, with the best results at epoch 28. The training duration was 1.789 hours. A batch size of 16 and an image size of 640 × 640 pixels were used. The YOLOv8x model consisted of 295 layers, 71,722,582 parameters, 0 gradients, and 343.7 GFLOPs.

\subsection{Performance Evaluation Metrics of YOLO Models for Strawberry Instance Segmentation}
The segmentation and performance of the YOLOv8 model configurations were evaluated using masked mean average precision (mAP), recall (R), and F1-score. Precision evaluates the accuracy of the predicted positive detections, calculated as: 
\[
\text{Precision} = \frac{\text{TP}}{\text{TP} + \text{FP}}
\]
Where:
\begin{itemize}
    \item TP = True Positives
    \item FP = False Positives
\end{itemize}

TP counts the correctly identified developing and maturing strawberries. FP indicating non-strawberries are incorrectly identified as ripe and unripe strawberries. FN denotes missing ripe and unripe strawberries. The area bounded by the recall rate, precision rate, and horizontal axis is measured by AP, which offers a measure of the detection model's performance across various threshold levels. Meanwhile, mAP is a single, combined performance metric that sums up the model's entire performance in detection. Averaging the Average Precision (AP) over all classes offers a comprehensive picture of the model's performance in object detection instances. Recall is a performance metric that indicates how many of the actual positives our model can identify, and it is computed as:
\[
\text{Recall} = \frac{\text{TP}}{\text{TP} + \text{FN}}
\]
Where:
\begin{itemize}
    \item TP = True Positives
    \item FN = False Negatives
\end{itemize}

The F1-score measures both the precision and recall of the model to compute a single score that represents the model's performance and is calculated as:
\[
\text{F1-Score} = \frac{2 \times (\text{Precision} \times \text{Recall})}{\text{Precision} + \text{Recall}}
\]

\subsection{ Counts of strawberry development detections }
Following the successful segmentation of ripe and unripe strawberry fruits using the YOLOv8 configuration models, the next step involved estimating detection counts per class. This step is crucial for numerous agricultural applications such as growth monitoring, yield prediction, and robotic crop management. To achieve this, detection counts were recorded per class after running inference on a test dataset consisting of 86 images. The dataset included 251 ground truth labels for ripe strawberries and 97 ground truth labels for unripe strawberries, totaling 348 ground truth labels.

\begin{table}[ht]
    \centering
    \caption{Total Detections per YOLOv8 Model Configuration}
    \label{tab:yolov8_detections}
    \begin{tabular}{cccc}
        \hline
        \textbf{Model} & \textbf{ Ripe Fruits} & \textbf{Unripe Fruits} & \textbf{Total Detections} \\
        \hline
        YOLOv8n & 235 & 51 & 286 \\
        YOLOv8s & 204 & 37 & 241 \\
        YOLOv8m & 221 & 51 & 272 \\
        YOLOv8l & 212 & 45 & 257 \\
        YOLOv8x & 224 & 50 & 274 \\
        \hline
    \end{tabular}
\end{table}
\section{Results and Discussion}
A total of 86 test images featuring ripe and unripe strawberries in the open field were utilized to evaluate the performance of various YOLOv8 model configurations in accurately segmenting these fruits. Table 2 summarizes the performance metrics for all the models after training. Among the five YOLOv8 model configurations tested, the YOLOv8n model demonstrated the highest performance, achieving a mAP@50 of 80.9\%. The YOLOv8n model also recorded the highest Precision at 80.2\%. Additionally, it achieved an F1 score of 78\% and a Recall of 77.3\%. A detailed breakdown of the performance metrics for all YOLOv8 model configurations is provided in Table 2 below.

\begin{table}[H]
    \centering
    \resizebox{\textwidth}{!}{ 
        \begin{tabular}{ccccc}
            \hline
            \textbf{Model} & \textbf{Precision} & \textbf{mAP@50} & \textbf{Recall} & \textbf{F1-score} \\
            \hline
            YOLOv8n & 0.802 & 0.809 & 0.773 & 0.780 \\
            YOLOv8s & 0.812 & 0.800 & 0.736 & 0.770 \\
            YOLOv8m & 0.813 & 0.784 & 0.731 & 0.770 \\
            YOLOv8l & 0.785 & 0.807 & 0.781 & 0.780 \\
            YOLOv8x & 0.803 & 0.785 & 0.719 & 0.760 \\
            \hline
        \end{tabular}
    }
    \caption{Performance of YOLOv8 Model Configurations}
    \label{tab:YOLOv8_model_configurations_performance}
\end{table}

Figures 6 and 7 display examples of ground truth images. Figures 8 through 12 present the results of predictions made using the five YOLOv8 model configurations. Table 3 provides the recorded pre-process, inference, and post-process speeds, measured in milliseconds (ms), for each YOLOv8 model—YOLOv8n, YOLOv8s, YOLOv8m, YOLOv8l, and YOLOv8x after running predictions on the test images.

 \begin{figure*}[h!] 
    \centering
    \begin{minipage}{0.49\textwidth}
        \centering
        \includegraphics[width=0.65\linewidth]{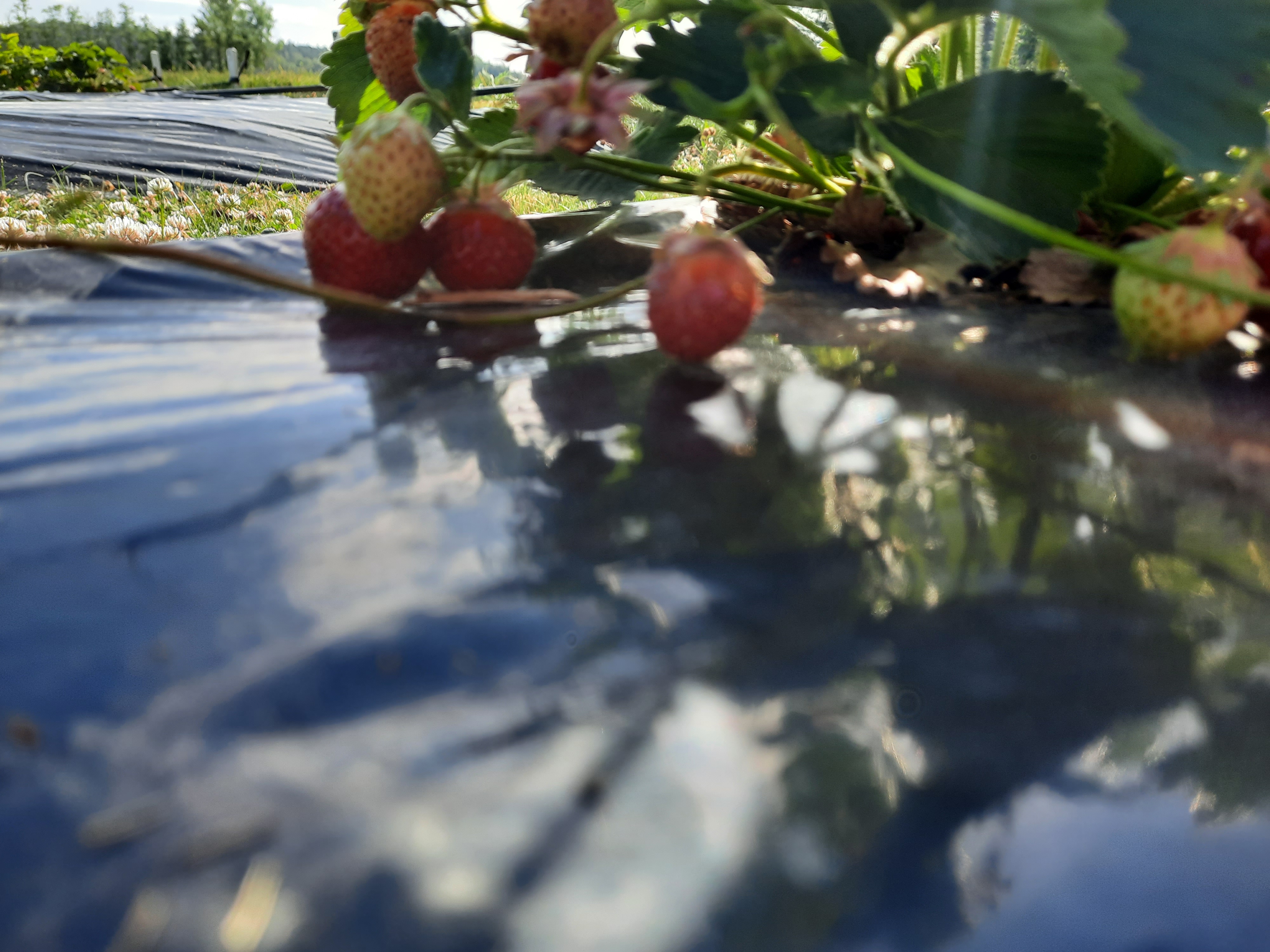}
        \caption{Ground truth image (a): Ripe and unripe strawberries}
        \label{fig:ground_truth_image_a}
    \end{minipage}
    \hfill
    \begin{minipage}{0.49\textwidth}
        \centering
        \includegraphics[width=0.65\linewidth]{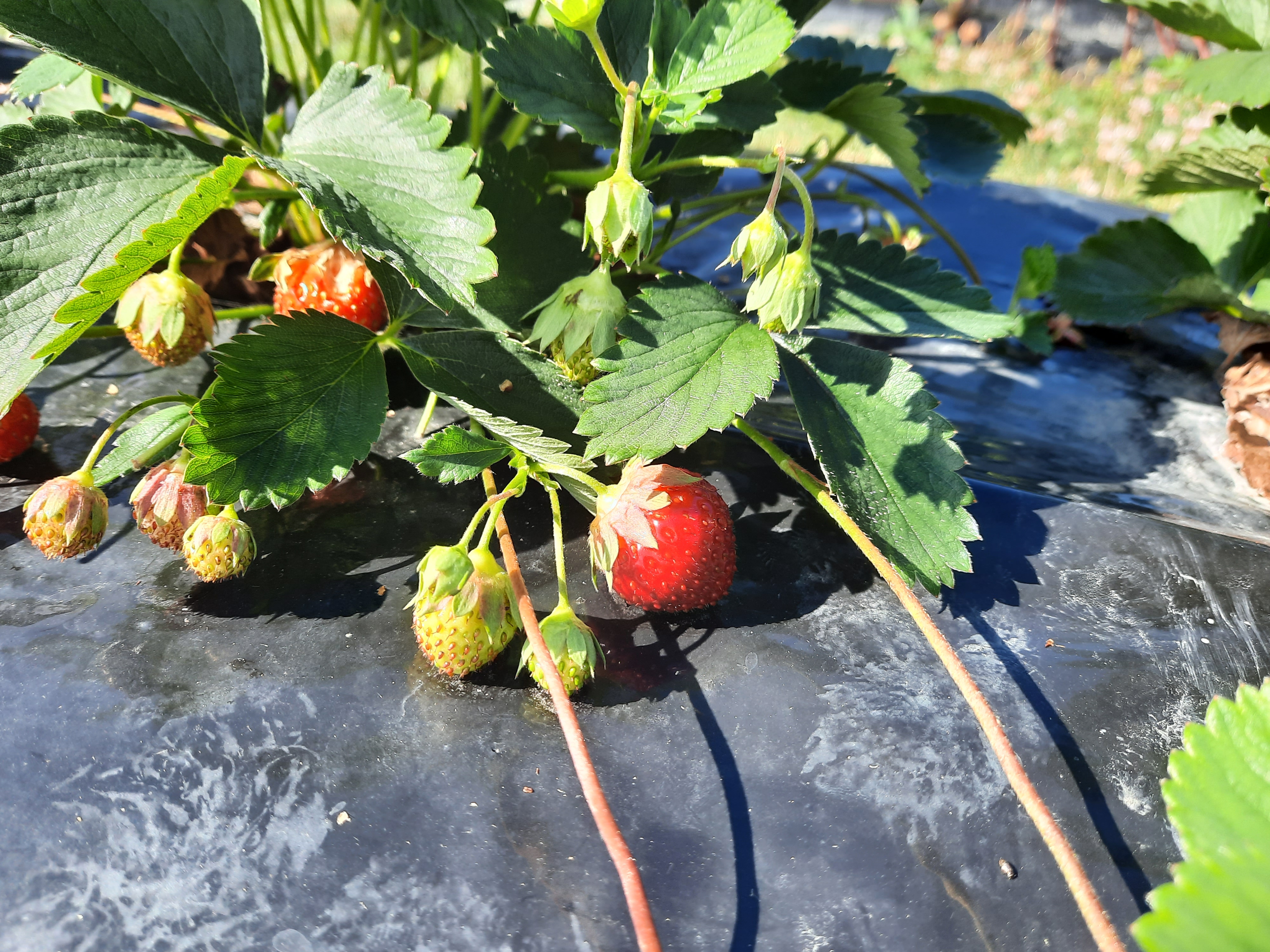}
        \caption{Ground truth image (b): Focus on unripe strawberries}
        \label{fig:ground_truth_image_b}
    \end{minipage}
\end{figure*}

\begin{figure*}[h!]
    \centering
    \begin{minipage}{0.49\textwidth}
        \centering
        \includegraphics[width=0.6\linewidth]{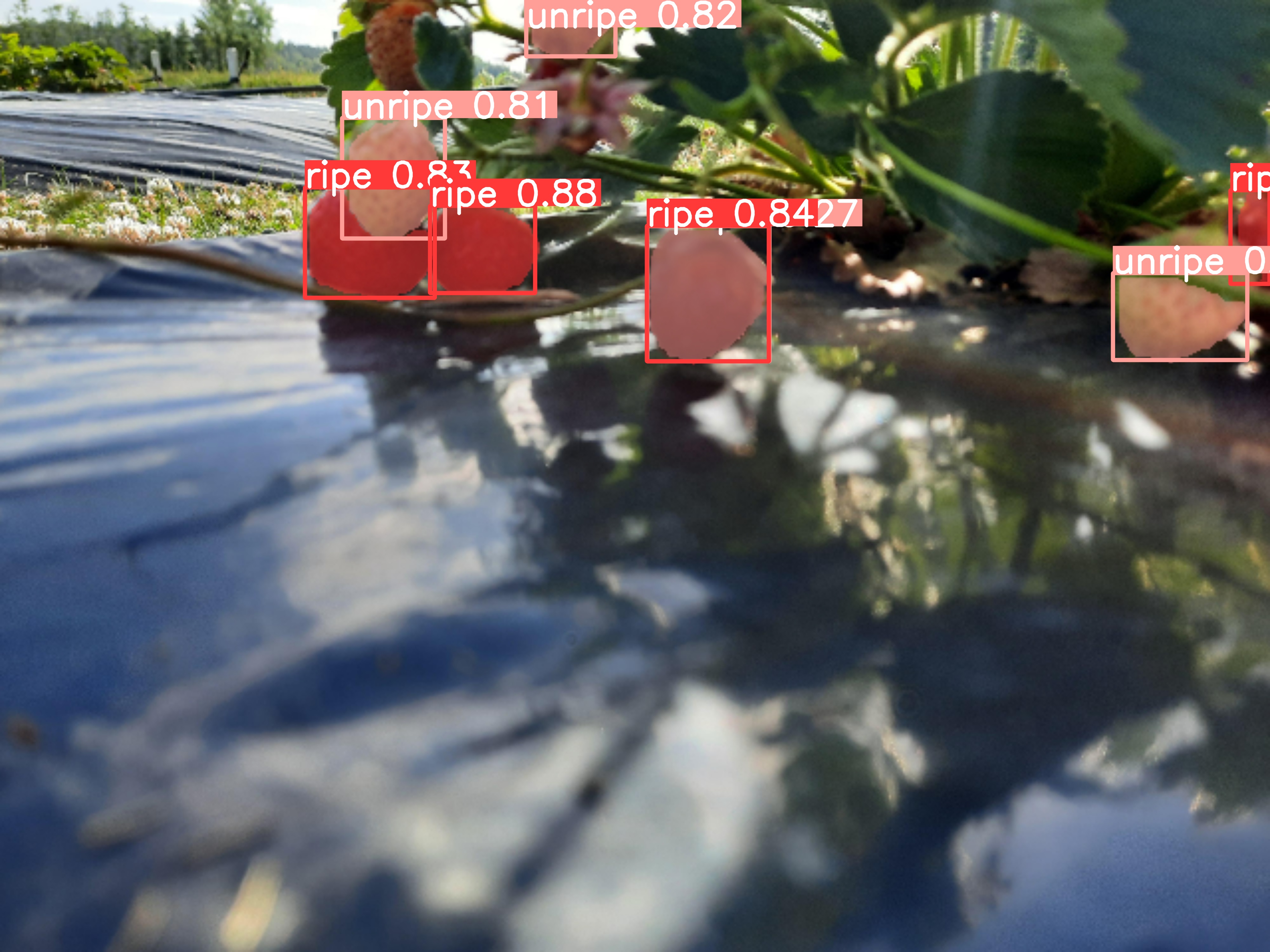}
        \caption{YOLOv8n (a): left}
        \label{fig:YOLOv8n_a_left}
    \end{minipage}
    \hfill
    \begin{minipage}{0.49\textwidth}
        \centering
        \includegraphics[width=0.65\linewidth]{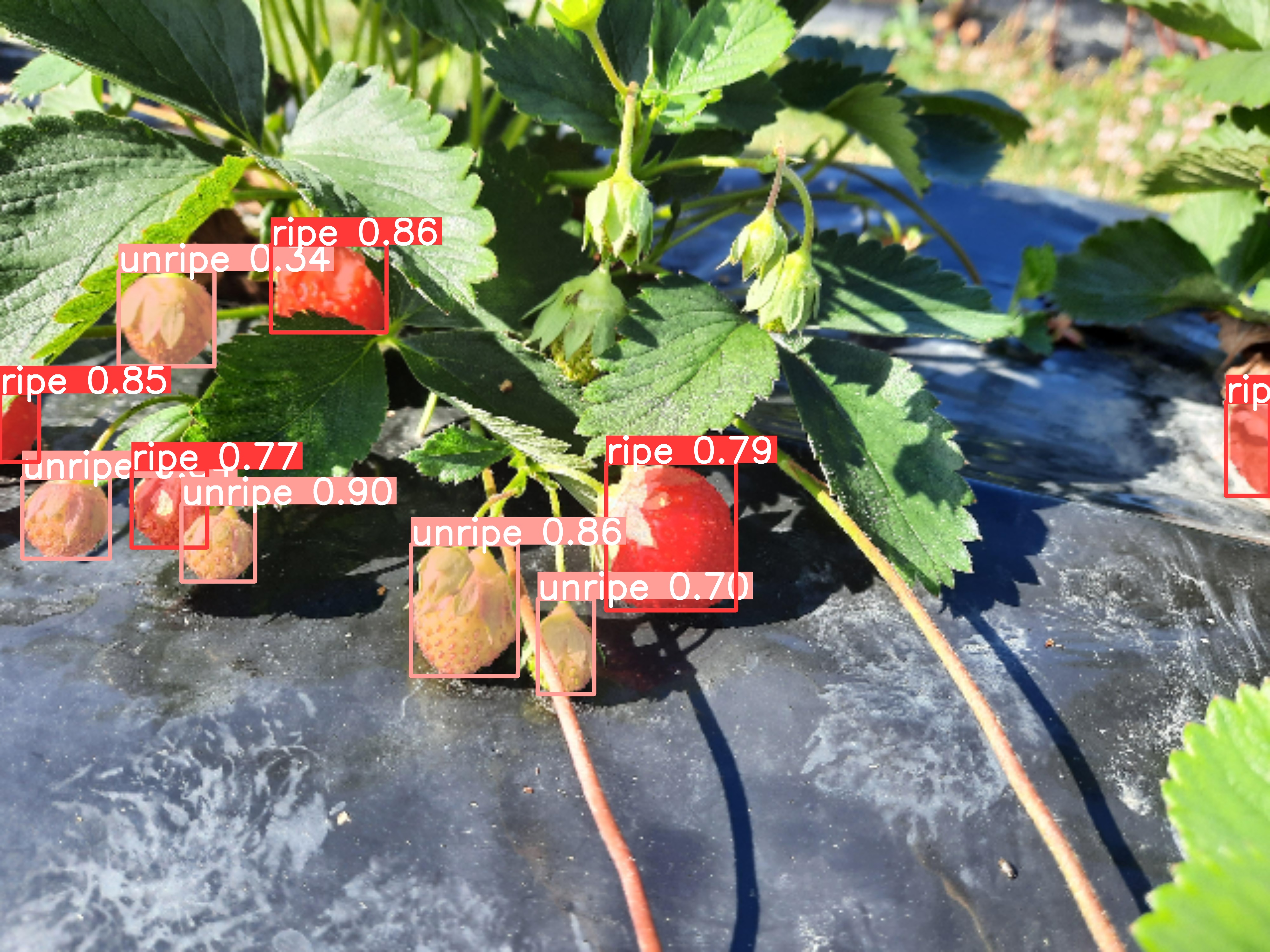}
        \caption{YOLOv8n (b): right}
        \label{fig:YOLOv8n_b_right}
    \end{minipage}
\end{figure*}

\begin{figure*}[h!]
    \centering
    \begin{minipage}{0.49\textwidth}
        \centering
        \includegraphics[width=0.65\linewidth]{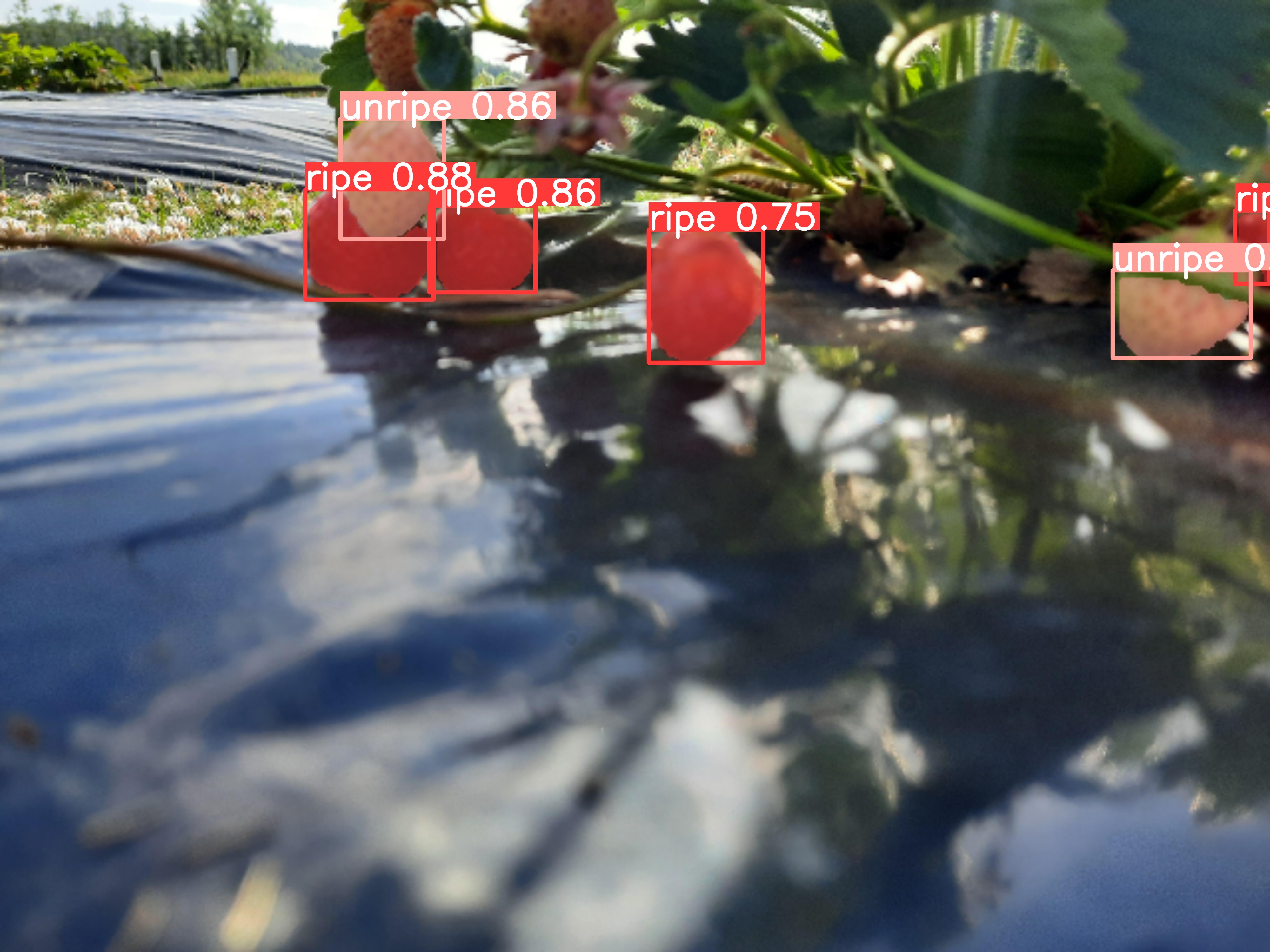}
        \caption{YOLOv8s (a): left}
        \label{fig:YOLOv8s_a_left}
    \end{minipage}
    \hfill
    \begin{minipage}{0.49\textwidth}
        \centering
        \includegraphics[width=0.6\linewidth]{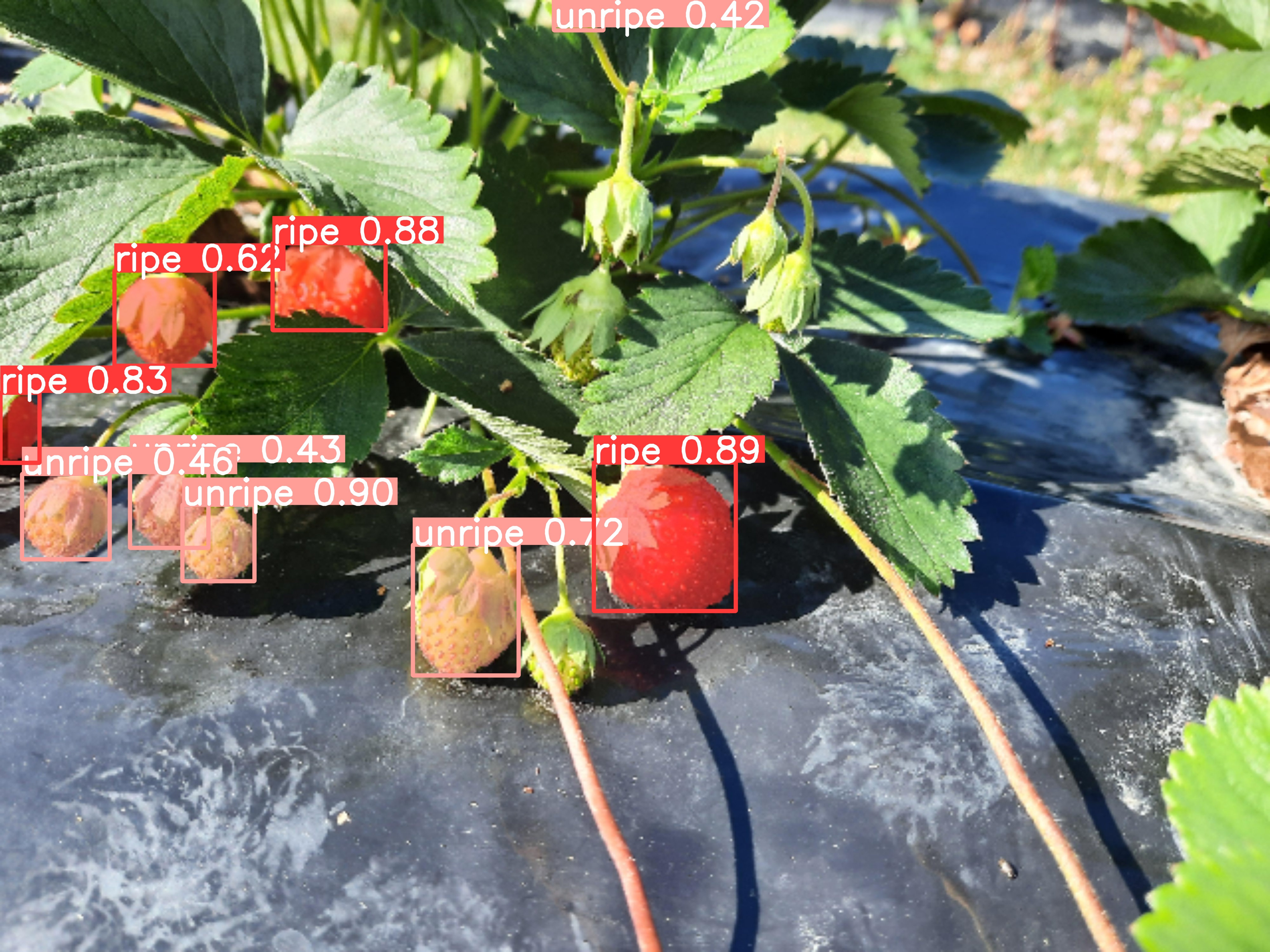}
        \caption{YOLOv8s (b): right}
        \label{fig:YOLOv8s_b_right}
    \end{minipage}
\end{figure*}

\begin{figure*}[h!]
    \centering
    \begin{minipage}{0.49\textwidth}
        \centering
        \includegraphics[width=0.65\linewidth]{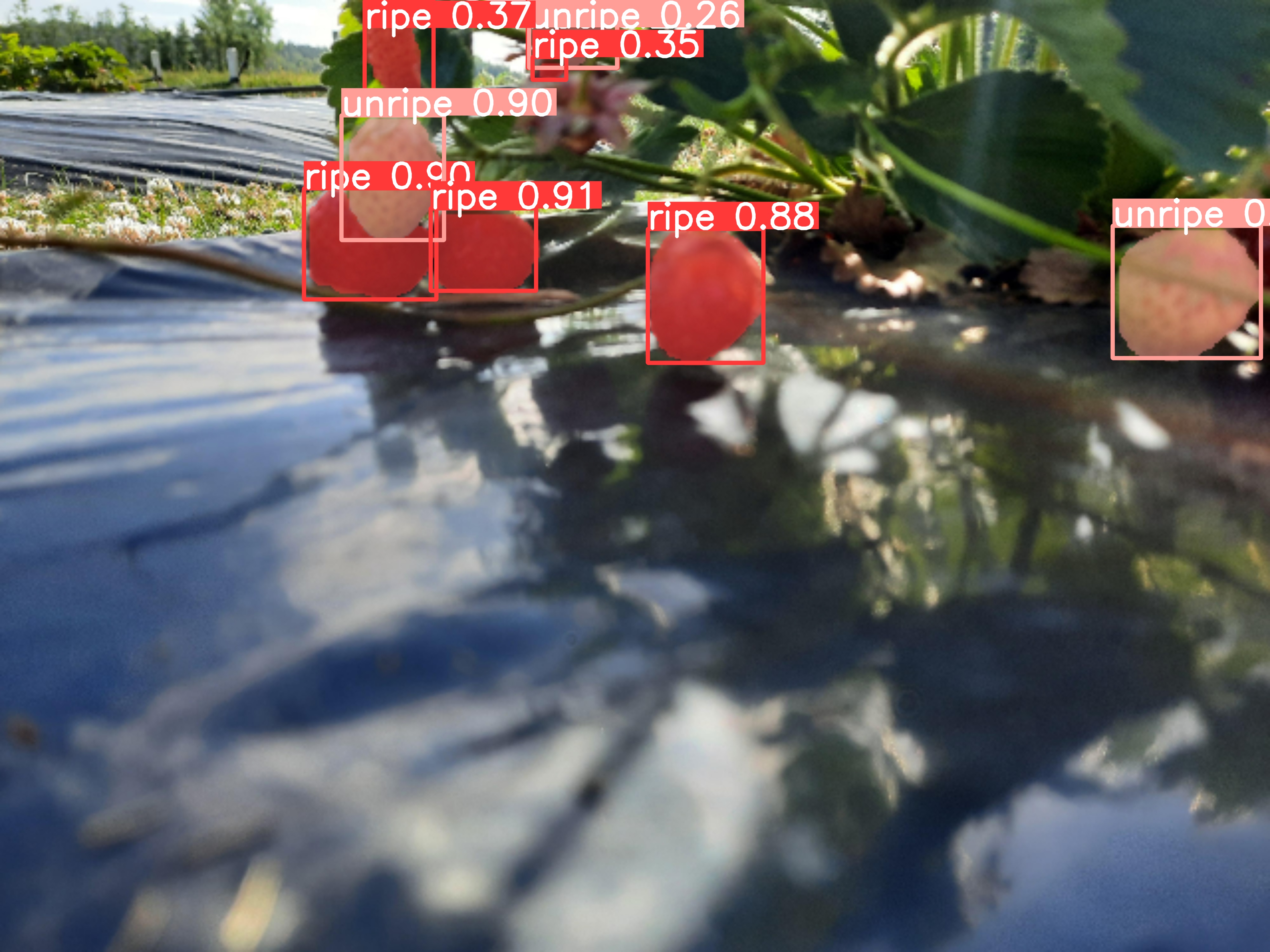}
        \caption{YOLOv8m (a): left}
        \label{fig:YOLOv8m_a_left}
    \end{minipage}
    \hfill
    \begin{minipage}{0.49\textwidth}
        \centering
        \includegraphics[width=0.65\linewidth]{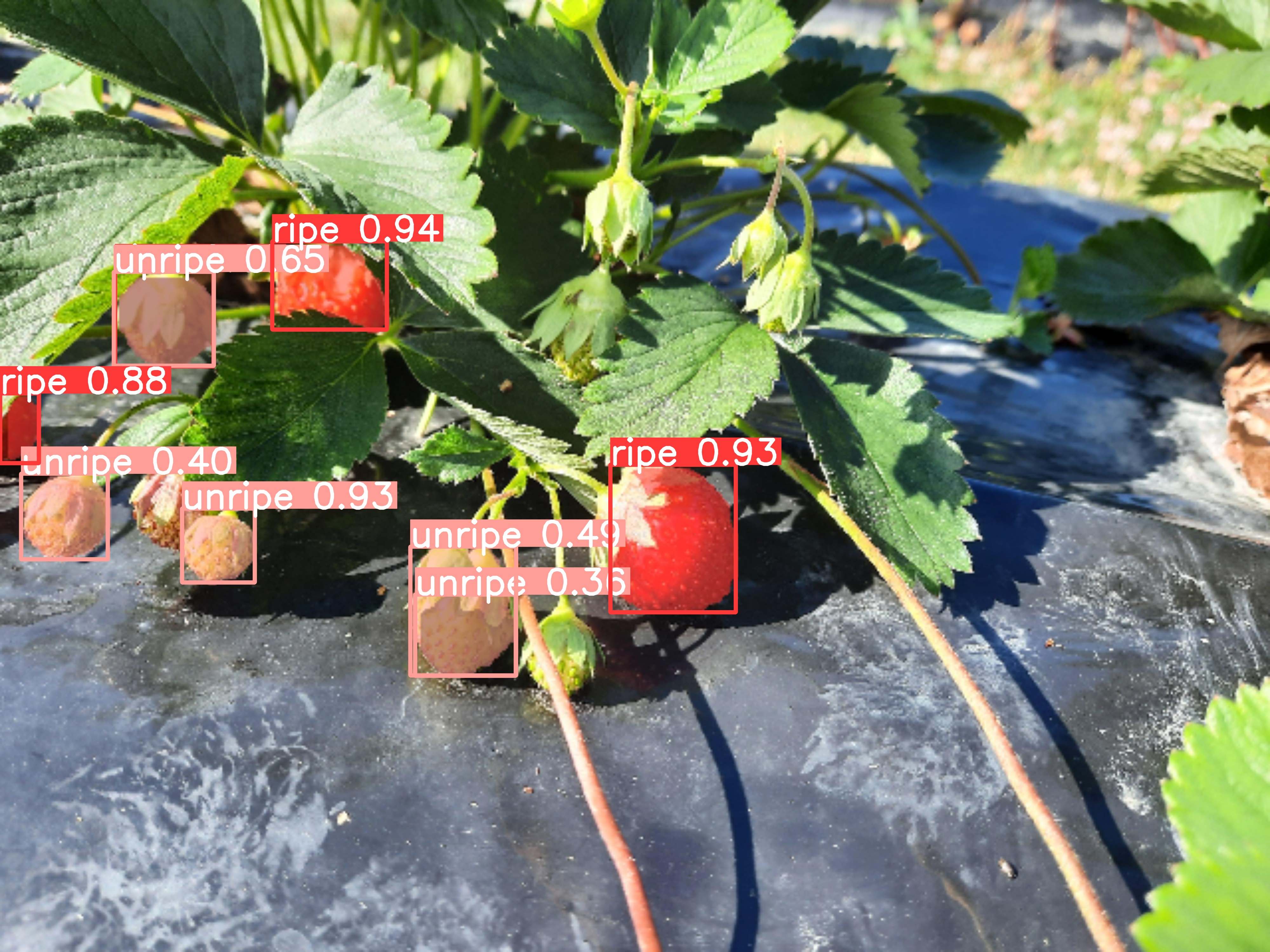}
        \caption{YOLOv8m (b): right}
        \label{fig:YOLOv8m_b_right}
    \end{minipage}
\end{figure*}

\begin{figure*}[h!]
    \centering
    \begin{minipage}{0.49\textwidth}
        \centering
        \includegraphics[width=0.65\linewidth]{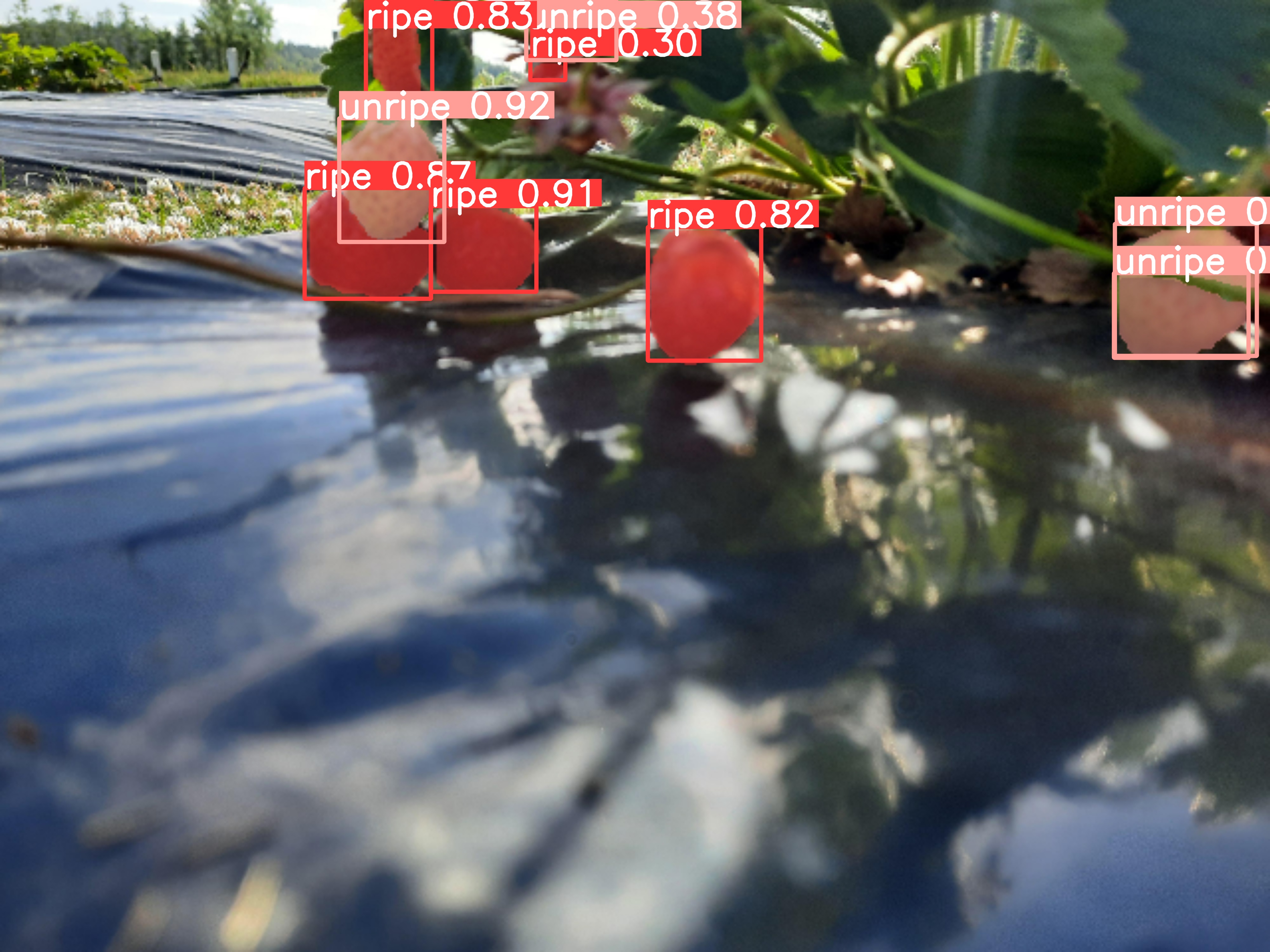}
        \caption{YOLOv8l (a): left}
        \label{fig:YOLOv8l_a_left}
    \end{minipage}
    \hfill
    \begin{minipage}{0.49\textwidth}
        \centering
        \includegraphics[width=0.65\linewidth]{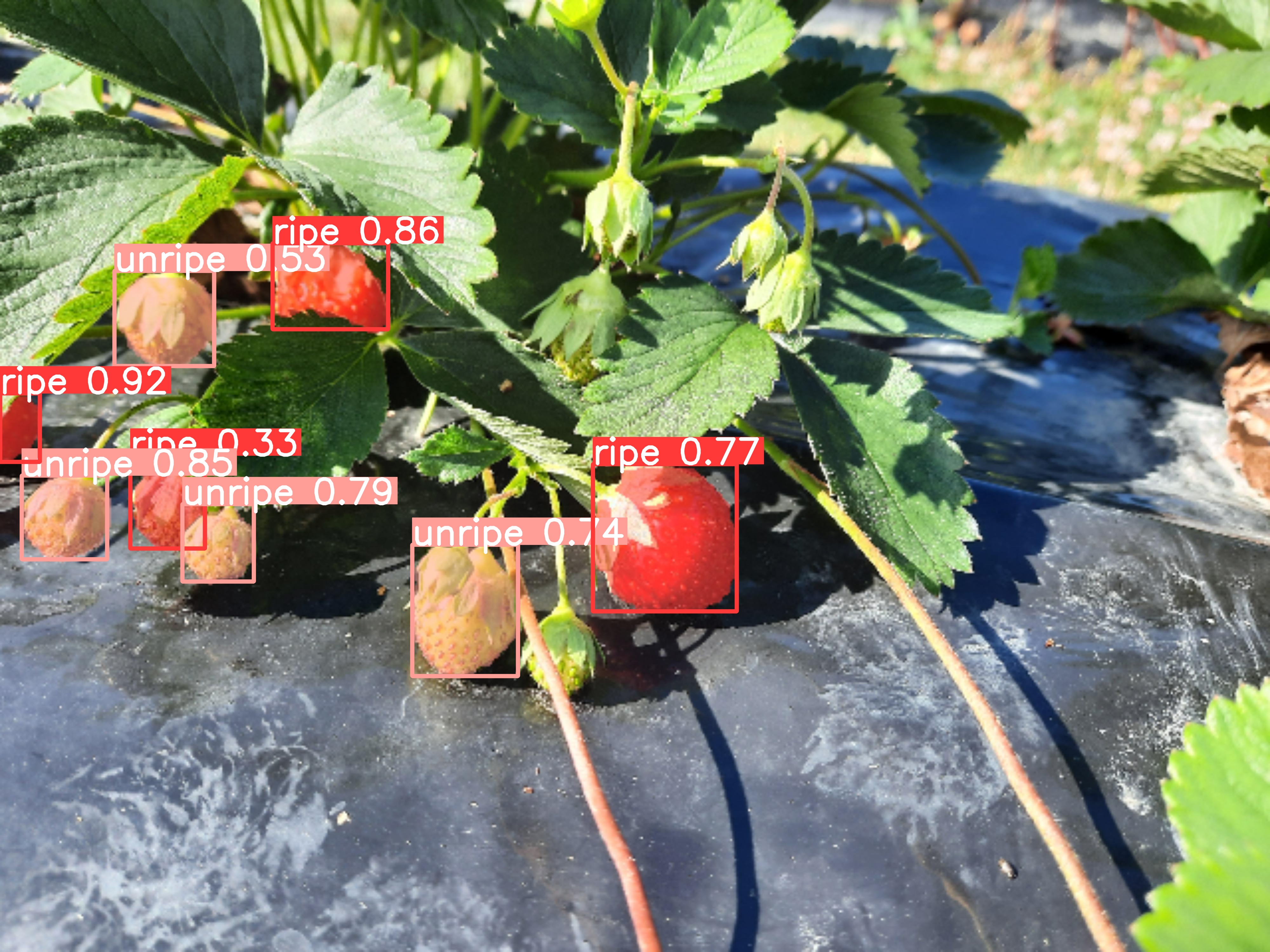}
        \caption{YOLOv8l (b): right}
        \label{fig:YOLOv8l_b_right}
    \end{minipage}
\end{figure*} 

\begin{table}[H]
    \centering
    \footnotesize
    \begin{tabular}{ccccc}
        \hline
        \textbf{Model} & \textbf{Pre-process (ms)} & \textbf{Inference (ms)} & \textbf{Post-process (ms)} & \textbf{Overall Inference (ms)} \\
        \hline
        YOLOv8n & 2.1 & 12.9 & 9.2 & 24.2 \\
        YOLOv8s & 2.2 & 22.2 & 8.6 & 33.0 \\
        YOLOv8m & 2.2 & 33.7 & 8.4 & 44.3 \\
        YOLOv8l & 2.3 & 42.8 & 8.5 & 53.6 \\
        YOLOv8x & 2.3 & 51.4 & 8.8 & 62.5 \\
        \hline
    \end{tabular}
    \caption{YOLOv8 Model Configurations Inference Speeds}
    \label{tab:YOLOv8_model_configurations_inference_speeds}
\end{table}

The YOLOv8n model, demonstrating the most satisfactory performance metrics, was selected for applications requiring optimal fruit detection and segmentation. This model not only achieved the highest performance but also exhibited the lowest overall inference speeds. The YOLOv8n model effectively detected both ripe and unripe strawberry fruits in an open-field setting. This study focused on comparing the performance of various models for segmenting ripe and unripe strawberries to identify the most optimal configuration for this application. The Recall-Confidence curve, F1-score curve, and Precision-Recall curve for the YOLOv8n model are presented in Figure 16 (a), (b), and (c).
\subsubsection{YOLOv8n:}
From the Precision-Recall curve, the model achieved a mean Average Precision (mAP) of 80.9\% across all classes with an Intersection over Union (IoU) threshold of 0.5. The Recall-Confidence curve recorded a recall of 93\% at the lowest confidence threshold. Regarding the F1-Confidence score curve, the model demonstrated an average precision of 78\% at a confidence threshold of 0.524.

\begin{figure}[ht]
    \centering
    \begin{subfigure}{0.32\textwidth}
        \centering
        \includegraphics[width=1.1\linewidth]{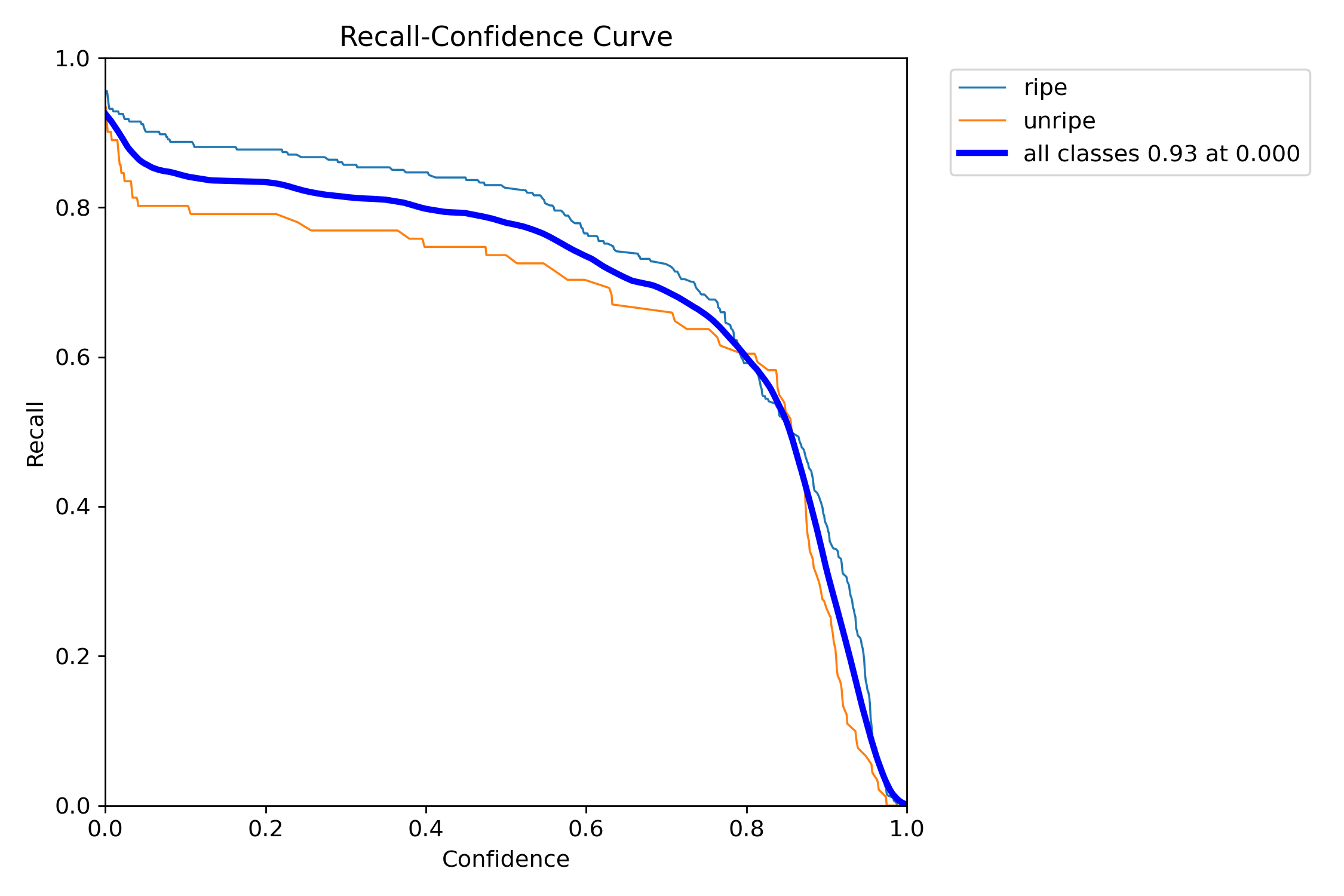}
        \caption{Recall Curve}
        \label{fig:Mask_Recall_curve}
    \end{subfigure}
    \hfill
    \begin{subfigure}{0.32\textwidth}
        \centering
        \includegraphics[width=1.1\linewidth]{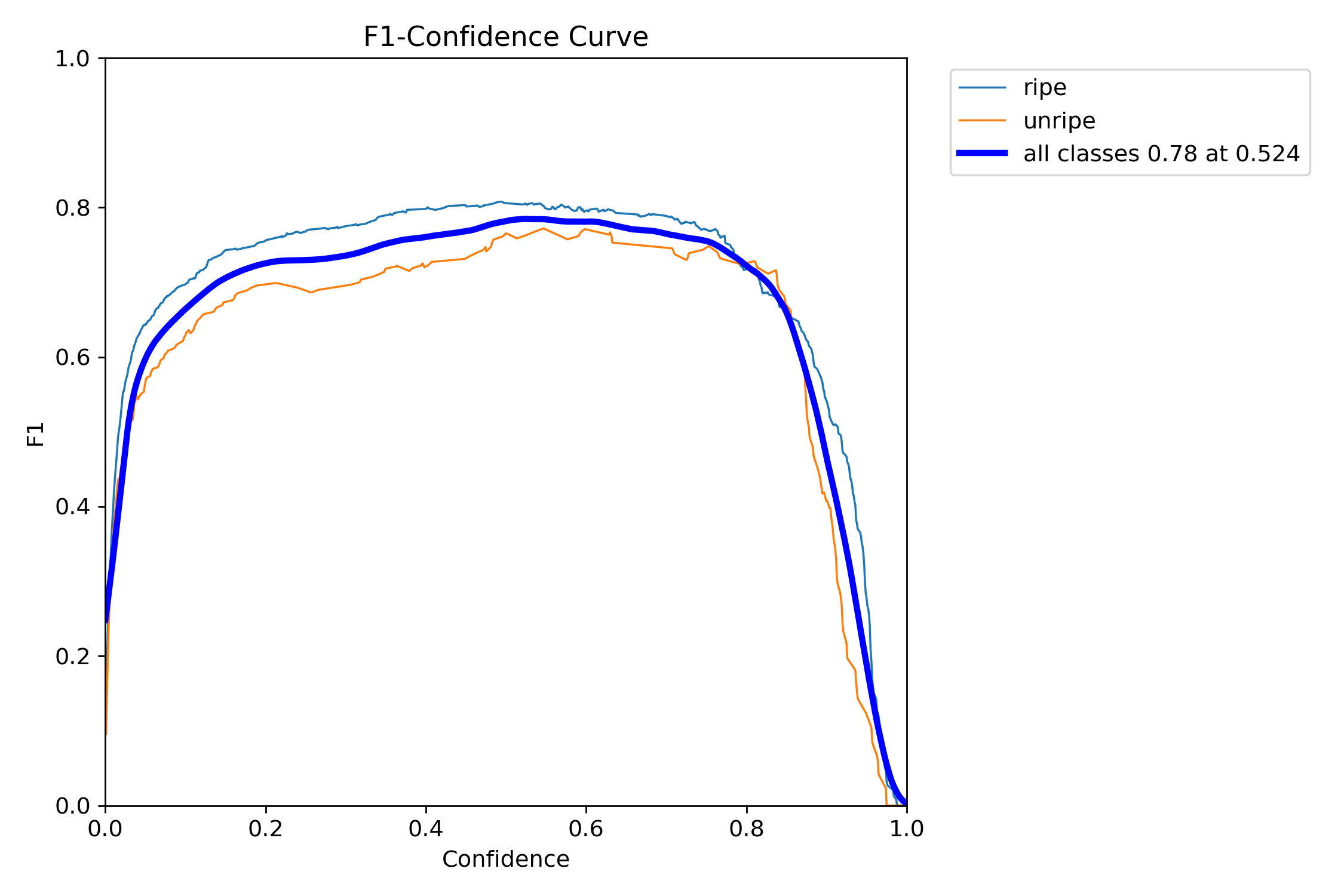}
        \caption{F1-Score Curve}
        \label{fig:Mask_F1_curve}
    \end{subfigure}
    \hfill
    \begin{subfigure}{0.32\textwidth}
        \centering
        \includegraphics[width=1.1\linewidth]{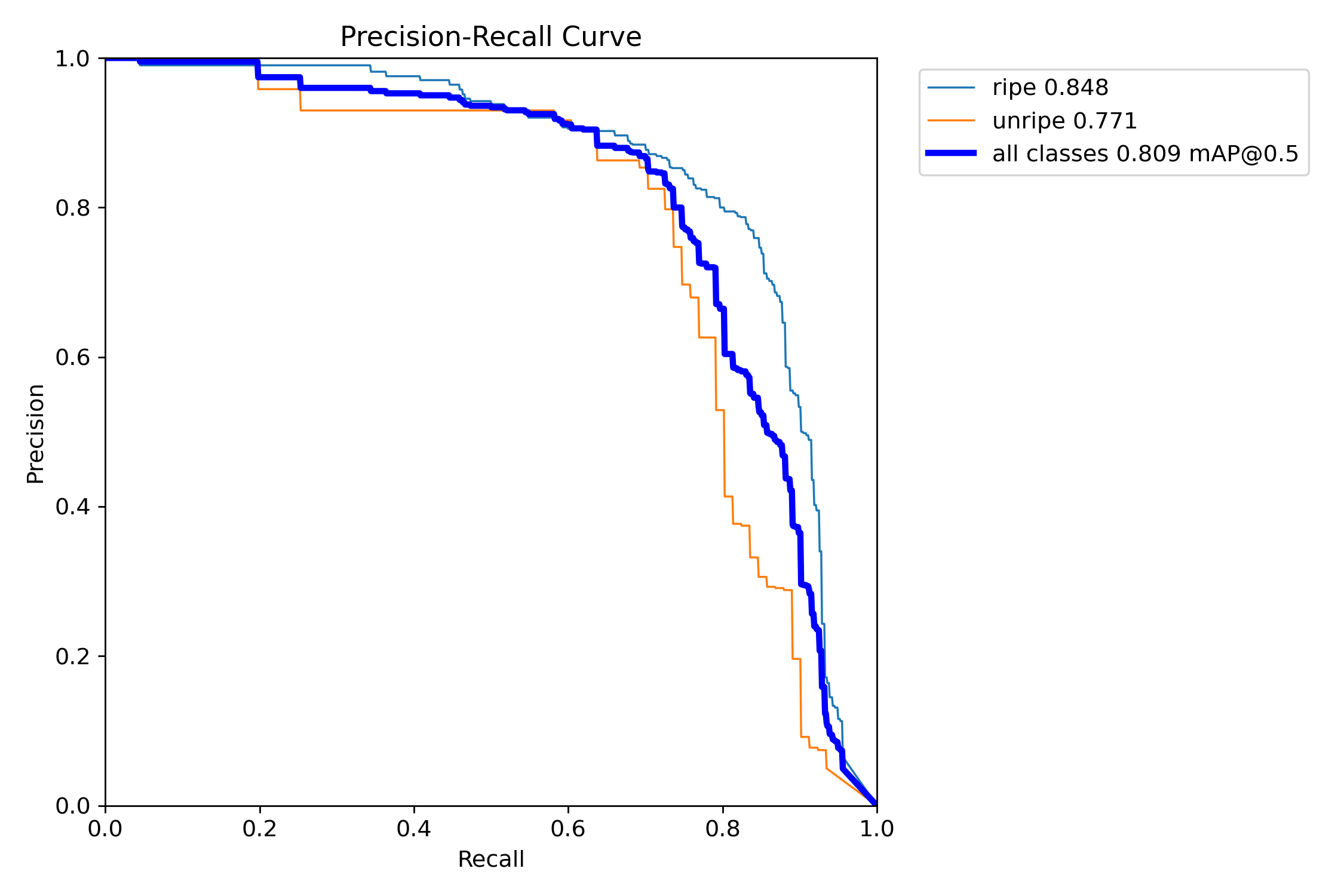}
        \caption{Precision-Recall Curve}
        \label{fig:Mask_Precision_Recall_curve}
    \end{subfigure}
    \caption{Performance Metrics for the Mask Model}
    \label{fig:Mask_performance_metrics}
\end{figure}

Recently, there has been a surge in research exploring YOLO-based algorithms for agricultural applications. Notably, \cite{wang2024real} focused on real-time detection and instance segmentation of strawberries in an unstructured Environment. \cite{sapkota2024immature} employed the YOLOv8 model for detecting and sizing immature apples in a commercial orchard. YOLOv5-based models have also been applied to other tasks, such as yield estimation for litchi fruits by \cite{wang2022fast} and accurate detection of green peppers by \cite{nan2023faster}. Additionally, \cite{wang2023improved} proposed an improved method for apple fruit target detection using the YOLOv5s model.

\section{Conclusion}
Instance segmentation of ripe and unripe strawberry fruits during their growth stages is crucial for various agricultural applications, including yield prediction, market strategizing, crop health assessment, and robotic fruit harvesting. While manual harvesting is effective, it is labor-intensive and time-consuming. Therefore, there is an urgent need for automation in horticultural processes such as harvesting. To address these challenges, this study focused on instance segmentation and estimating the quantity of detected ripe and unripe strawberries using the YOLOv8 model for yield estimation and growth stage monitoring. The major findings of this study include: 
\begin{itemize}
    \item The YOLOv8n instance segmentation model demonstrated superior performance in detecting both ripe and unripe strawberry fruits, achieving a mean Average Precision (mAP) of 80.9\%. It consistently outperformed the other models tested on the same dataset, accurately detecting 286 out of 348 strawberry fruits.

\end{itemize}
\begin{itemize}
    \item The YOLOv8n model also achieved the fastest overall inference speed of 24.2 ms compared to the other YOLOv8 model configurations. Automating the detection and yield estimation of ripe and unripe fruits not only addresses the labor-intensive and time-consuming nature of manual harvesting but also reduces overall costs while improving crop health and quality. For future research, expanding the dataset and capturing images with advanced agricultural cameras mounted on manned or unmanned ground vehicles would enhance real-life data collection by agricultural robots in the field. Additionally, exploring other advanced machine-learning algorithms could help minimize the impact of variable open field conditions and occlusions caused by leaves, branches, and other fruits. Shortly, the results of this study could be applied in strawberry fields, utilizing the YOLOv8n model to accurately detect and estimate the yield of strawberry fruit development. For automating the harvesting of mature strawberries, integrating embedded computational hardware and software with a robotic system would be necessary to implement the YOLOv8n model effectively.
\end{itemize}

\begin{itemize}
    \item Additionally, for early yield prediction, a cost-effective sensing module could be developed using RGB-D cameras and an image acquisition interface. This module would collect images that could be processed using either in-house or cloud computing platforms. The results could then be presented to growers through user-friendly software interfaces, which could be accessed via web or mobile applications, enabling informed decision-making.
\end{itemize}

\section{Acknowledgment}
The authors express their profound gratitude to the creators of the dataset \cite{pastell_matti_2022_6126677} used in this study. We also extend our thanks to Dr. Magni Stentoft Hussain for his invaluable advice throughout the research.

\bibliography{library}

\begin{thebibliography}{10}

\bibitem{Mittal2019}
U.~Mittal, S.~Srivastava, and P.~Chawla, ``Review of different techniques for object detection using deep learning,'' {\em ACM International Conference Proceeding Series}, 6 2019.

\bibitem{Tian2020}
H.~Tian, T.~Wang, Y.~Liu, X.~Qiao, and Y.~Li, ``Computer vision technology in agricultural automation —a review,'' {\em Information Processing in Agriculture}, vol.~7, pp.~1--19, 3 2020.

\bibitem{Delnevo2022}
G.~Delnevo, R.~Girau, C.~Ceccarini, and C.~Prandi, ``A deep learning and social iot approach for plants disease prediction toward a sustainable agriculture,'' {\em IEEE Internet of Things Journal}, vol.~9, pp.~7243--7250, 5 2022.

\bibitem{Sharma2021}
A.~Sharma, A.~Jain, P.~Gupta, and V.~Chowdary, ``Machine learning applications for precision agriculture: A comprehensive review,'' 2021.

\bibitem{giampieri2012strawberry}
F.~Giampieri, S.~Tulipani, J.~M. Alvarez-Suarez, J.~L. Quiles, B.~Mezzetti, and M.~Battino, ``The strawberry: Composition, nutritional quality, and impact on human health,'' {\em Nutrition}, vol.~28, no.~1, pp.~9--19, 2012.

\bibitem{li2011review}
P.~Li, S.-h. Lee, and H.-Y. Hsu, ``Review on fruit harvesting method for potential use of automatic fruit harvesting systems,'' {\em Procedia Engineering}, vol.~23, pp.~351--366, 2011.

\bibitem{Stajnko2004}
D.~Stajnko, M.~Lakota, and M.~Hočevar, ``Estimation of number and diameter of apple fruits in an orchard during the growing season by thermal imaging,'' {\em Computers and Electronics in Agriculture}, vol.~42, pp.~31--42, 2004.

\bibitem{pest2024}
``Pests of fruit crops: A colour handbook, second edition - david v alford - google books.''

\bibitem{Campos2021}
C.~N.~S. Campos, G.~C.~M. Teixeira, R.~de~Mello~Prado, G.~Caione, G.~B. da~Silva~Júnior, C.~H.~O. de~David, A.~C. Sales, C.~G. Roque, and P.~E. Teodoro, ``Macronutrient deficiency in cucumber plants: impacts in nutrition, growth and symptoms,'' {\em Journal of Plant Nutrition}, vol.~44, pp.~2609--2626, 10 2021.

\bibitem{Wu2019}
J.~Wu, B.~Zhang, J.~Zhou, Y.~Xiong, B.~Gu, and X.~Yang, ``Automatic recognition of ripening tomatoes by combining multi-feature fusion with a bi-layer classification strategy for harvesting robots,'' {\em Sensors 2019, Vol. 19, Page 612}, vol.~19, p.~612, 2 2019.

\bibitem{Erfani2019}
S.~Erfani, A.~Jafari, and A.~Hajiahmad, ``Comparison of two data fusion methods for localization of wheeled mobile robot in farm conditions,'' {\em Artificial Intelligence in Agriculture}, vol.~1, pp.~48--55, 3 2019.

\bibitem{yuan2011greenhouse}
T.~Yuan, C.~Ji, Y.~Chen, W.~Li, and J.~Zhang, ``Greenhouse cucumber recognition based on spectral imaging technology,'' {\em Trans. Chin. Soc. Agric. Mach}, vol.~42, pp.~172--176, 2011.

\bibitem{Liu2020}
G.~Liu, J.~C. Nouaze, P.~L.~T. Mbouembe, and J.~H. Kim, ``Yolo-tomato: A robust algorithm for tomato detection based on yolov3,'' {\em Sensors (Switzerland)}, vol.~20, 4 2020.

\bibitem{Wang2021}
X.~Wang and J.~Liu, ``Tomato anomalies detection in greenhouse scenarios based on yolo-dense,'' {\em Frontiers in Plant Science}, vol.~12, p.~634103, 4 2021.

\bibitem{ruparelia2022real}
S.~Ruparelia, M.~Jethva, and R.~Gajjar, ``Real-time tomato detection, classification, and counting system using deep learning and embedded systems,'' in {\em Proceedings of the International e-Conference on Intelligent Systems and Signal Processing: e-ISSP 2020}, pp.~511--522, Springer, 2022.

\bibitem{Wang2023}
H.~. Wang, J.~. Feng, H.~Yin, Y.~Song, W.~Wen, H.~Wang, J.~Feng, and H.~Yin, ``Improved method for apple fruit target detection based on yolov5s,'' {\em Agriculture 2023, Vol. 13, Page 2167}, vol.~13, p.~2167, 11 2023.

\bibitem{Borrero2020}
I.~Pérez-Borrero, D.~Marín-Santos, M.~E. Gegúndez-Arias, and E.~Cortés-Ancos, ``A fast and accurate deep learning method for strawberry instance segmentation,'' {\em Computers and Electronics in Agriculture}, vol.~178, 11 2020.

\bibitem{Afzaal2021}
U.~Afzaal, B.~Bhattarai, Y.~R. Pandeya, and J.~Lee, ``An instance segmentation model for strawberry diseases based on mask r-cnn,'' {\em Sensors}, vol.~21, 10 2021.

\bibitem{Lawal2023}
O.~M. Lawal, ``Yolov5-linet: A lightweight network for fruits instance segmentation,'' {\em PLoS ONE}, vol.~18, 3 2023.

\bibitem{Yang2023}
G.~Yang, J.~Wang, Z.~Nie, H.~Yang, and S.~Yu, ``A lightweight yolov8 tomato detection algorithm combining feature enhancement and attention,'' {\em Agronomy}, vol.~13, 7 2023.

\bibitem{Jia2022}
W.~Jia, M.~Liu, R.~Luo, C.~Wang, N.~Pan, X.~Yang, and X.~Ge, ``Yolof-snake: An efficient segmentation model for green object fruit,'' {\em Frontiers in Plant Science}, vol.~13, 6 2022.

\bibitem{Chai2023}
J.~J. Chai, J.~L. Xu, and C.~O’Sullivan, ``Real-time detection of strawberry ripeness using augmented reality and deep learning,'' {\em Sensors}, vol.~23, 9 2023.

\bibitem{Li2024}
Y.~Li, W.~Wang, X.~Guo, X.~Wang, Y.~Liu, and D.~Wang, ``Recognition and positioning of strawberries based on improved yolov7 and rgb-d sensing,'' {\em Agriculture (Switzerland)}, vol.~14, 4 2024.

\bibitem{Yue2023}
X.~Yue, K.~Qi, X.~Na, Y.~Zhang, Y.~Liu, and C.~Liu, ``Improved yolov8-seg network for instance segmentation of healthy and diseased tomato plants in the growth stage,'' {\em Agriculture (Switzerland)}, vol.~13, 8 2023.

\bibitem{Yu2019}
Y.~Yu, K.~Zhang, L.~Yang, and D.~Zhang, ``Fruit detection for strawberry harvesting robot in non-structural environment based on mask-rcnn,'' {\em Computers and Electronics in Agriculture}, vol.~163, 8 2019.

\bibitem{Yamamoto}
S.~Yamamoto, S.~Hayashi, S.~Saito, Y.~Ochiai, T.~Yamashita, and S.~Sugano, ``Development of robotic strawberry harvester to approach target fruit from hanging bench side.''

\bibitem{Gonzalez2019}
S.~Gonzalez, C.~Arellano, and J.~E. Tapia, ``Deepblueberry: Quantification of blueberries in the wild using instance segmentation,'' {\em IEEE Access}, vol.~7, pp.~105776--105788, 2019.

\bibitem{lemsalu2022real}
M.~Lemsalu, V.~Bloch, J.~Backman, and M.~Pastell, ``Real-time cnn-based computer vision system for open-field strawberry harvesting robot,'' {\em IFAC-PapersOnLine}, vol.~55, no.~32, pp.~24--29, 2022.

\bibitem{wang2024real}
C.~Wang, F.~Ding, Y.~Wang, R.~Wu, X.~Yao, C.~Jiang, and L.~Ling, ``Real-time detection and instance segmentation of strawberry in unstructured environment.,'' {\em Computers, Materials \& Continua}, vol.~78, no.~1, 2024.

\bibitem{sapkota2024immature}
R.~Sapkota, D.~Ahmed, M.~Churuvija, and M.~Karkee, ``Immature green apple detection and sizing in commercial orchards using yolov8 and shape fitting techniques,'' {\em IEEE Access}, vol.~12, pp.~43436--43452, 2024.

\bibitem{wang2022fast}
L.~Wang, Y.~Zhao, Z.~Xiong, S.~Wang, Y.~Li, and Y.~Lan, ``Fast and precise detection of litchi fruits for yield estimation based on the improved yolov5 model,'' {\em Frontiers in Plant Science}, vol.~13, p.~965425, 2022.

\bibitem{nan2023faster}
Y.~Nan, H.~Zhang, Y.~Zeng, J.~Zheng, and Y.~Ge, ``Faster and accurate green pepper detection using nsga-ii-based pruned yolov5l in the field environment,'' {\em Computers and Electronics in Agriculture}, vol.~205, p.~107563, 2023.

\bibitem{wang2023improved}
H.~Wang, J.~Feng, and H.~Yin, ``Improved method for apple fruit target detection based on yolov5s,'' {\em Agriculture}, vol.~13, no.~11, p.~2167, 2023.

\bibitem{pastell_matti_2022_6126677}
L.~M. Pastell~Matti and B.~Victor, ``Strawberry dataset for object detection,'' Feb. 2022.

\end{thebibliography}
\FloatBarrier
\end{document}